\documentclass[journal]{IEEEtran}
\usepackage{blindtext}
\usepackage{graphicx}
\usepackage[fleqn]{amsmath}
\usepackage{bbold}
\usepackage{url}
\usepackage{epstopdf}

\usepackage[T1]{fontenc}
\usepackage[utf8x]{inputenc}

\ifCLASSINFOpdf

\else
 
\fi

\begin{document}

\title{Simulated Annealing Algorithm for Graph Coloring}

\author{Alper~Köse,
        Berke Aral~Sönmez, Metin~Balaban,
        ~\IEEEmembership{Random Walks Project}
       
}

\maketitle

\begin{abstract}
 The goal of this Random Walks project is to code and experiment the Markov Chain Monte Carlo (MCMC) method for the problem of graph coloring. In this report, we present the plots of cost function  \(\mathbf{H}\) by varying the parameters like \(\mathbf{q}\) (Number of colors that can be used in coloring) and \(\mathbf{c}\) (Average node degree). The results are obtained by using simulated annealing scheme, where the temperature (inverse of \(\mathbf{\beta}\)) parameter in the MCMC is lowered progressively.
\end{abstract}


\IEEEpeerreviewmaketitle

\section{Introduction}

\subsection{Graph Coloring}

Graph coloring is one of the most important concepts in graph theory and is used in many real time applications in computer science. It can be defined as a problem of how to assign colors to certain elements of a graph given some constraints. It is especially used in research areas of science such as data mining, image segmentation, clustering, image capturing, networking etc.
Vertex coloring is the most common application of graph coloring. In vertex coloring, given \(m\) colors, the aim is to find the coloring of vertices of the graph such that no two adjacent vertices have the same color (Here, adjacent means that two vertices are directly connected by an edge). If this goal is achieved, the graph is said to have proper coloring. However, it is not always possible to find the proper coloring of a graph. In this situation, the coloring with the minimum cost function is searched:

\begin{equation}
H(x)=\sum_{(v,w)\in E}  \mathbf{1}_{x_{v}=x_{w}}
\end{equation}


This energy function assigns a unit cost to every edge which has the same colors on two vertices bounded to it. Therefore, one can state that this function counts the total number of such bad edges in a graph. Our goal here is to minimize the energy function and to see if there exist minimizers that yield zero cost (proper coloring). On the other hand, if the global minimum is strictly positive, then there does not exist a proper coloring for the given graph for given number of colors.

To minimize the energy function, we use the simulated annealing method in this project.

\subsection{Simulated Annealing}

Finding the optimal solution for some optimization problems can be an incredibly difficult task because when a problem gets sufficiently large we need to search through an enormous number of possible solutions to find the optimal one. Therefore, there are often too many possible solutions to consider. In such cases, since it is not possible to find the optimal solution within a reasonable length of time, one has to try to find a solution that is close enough to global optimum.

Most minimization strategies find the nearest local minimum but we need a different strategy not to get stuck in a local minimum. Namely, we should sometimes iterate to new colorings that do not improve the current situation. In such a situation, we consider Simulated Annealing, since it is a probabilistic technique for approximating the global optimum of a given function which works as described above.

In Simulated Annealing, there is a temperature parameter that has to be tuned, in order to get close enough to the optimal solution. It has an initial value and it should be slowly decreased. One must be careful for finding such a schedule because if the cooling is too slow, it may not be possible to reach the minimizer in a reasonable amount of time, on the other hand, if the cooling is too fast, it is likely to end up in a local minimum.

For now, practical deterministic algorithms for finding optimal solution are not known. As a probabilistic model, Simulated Annealing method can do quite well. In other words, given sufficient time at each temperature, it leads to equilibrium state and in the end it can converge to a solution which is close enough to the optimal one.

\subsection{Metropolis Algorithm}

In the problem, we want to find the coloring vector that minimizes the cost function \(H\). In order to achieve this goal, we use simulated annealing which is an adaptation of Metropolis Algorithm and the Monte Carlo method. Metropolis Algorithm that is given in the project description is used in optimization with only modifying \(\mathbf{\beta}\) parameter in some steps of the iterations.

The Metropolis algorithm is given below (Think \(\mathbf{\beta}\) as fixed for now since we do not change it in every iteration) :

1. At time \(t = 0\), initialize with a coloring  \(x^{0}\) taken uniformly at random.

2. At times \(t \geq 0\), make a transition \(x^{t} \rightarrow x^{t+1}\) according to the following rules:

\(\bullet\) Select a vertex \(v \in V\) uniformly at random, with current color \(x_{v}\). Pick a color different from \(x_{v}\) at random and recolor the vertex  \(v\). Consider the new coloring \(x^{new}\) and compute \(\bigtriangleup = H(x^{new})-H(x^{t})\).

\(\bullet\) if \(\bigtriangleup \leq 0\), then accept the new color (with probability one) and set \(x^{t+1} = x^{new}\).

\(\bullet\) if \(\bigtriangleup>0\), then accept the new color with probability
\begin{equation}
\frac {p(H(x^{new}))} {p(H(x^{t}))} = exp(-\beta\bigtriangleup)
\end{equation}

and set \(x^{t+1} = x^{new}\); reject the move with the complementary probability and set \(x^{t+1} = x^{t}\). 

3. Iterate for \(n\) time steps.

\begin{figure}[h!]
    \centering
    \scalebox{0.45}{\includegraphics{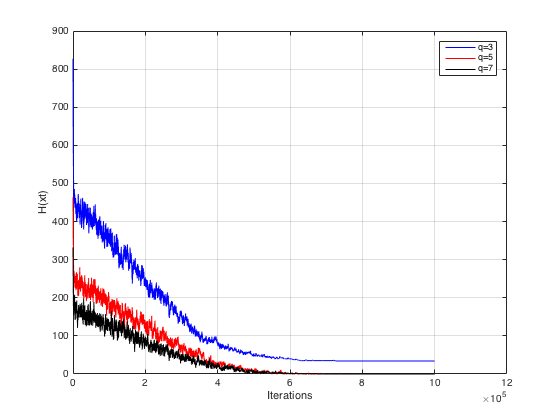}}
    \caption{Results on an Erdos-Renyi graph with $1000$ vertices and average node degree $c=5$. The algorithm is iterated $10^6$ times for every $q$ value of $3$,$5$ and $7$. For these $q$ values, $H_{min}$ is obtained as $34$, $0$, and $0$ respectively.}
\end{figure} 

\section{Experimental Setup and Discussion of Results}

As recommended in the lab sheet, we coded the algorithm in Matlab firstly. We know that there is not a certain optimal point that is aimed to be reached. We want to make \(H_{min}\) as small as possible. As known, \(\beta\) must be increased slow enough not to converge to a local minimum and fast enough to complete the simulation in a reasonable amount of time. By trials we saw that increasing the number of iterations allows us to reach lower energies and sometimes proper colorings for some parameters. Therefore, we chose our iteration number to be as big as possible and also as small as allowing as to do computations in a reasonable amount of time. We experimented with 1000x1000 Erdos-Renyi random graph to see if the execution time is reasonable. When we do \(10^{5}\) iterations, it takes approximately 50 seconds to run the program in Matlab. We thought that we can complete this task in a time which is much shorter than this, so we coded the algorithm in C++. As a result, it took 0.04 seconds to run the same algorithm which is 1250 times faster. 

Then, in the optimization step, we thought that we can increase \(\beta\) slowly not to get stuck in a local minima since our computation time is really low. With the trials by hand, we estimated the modification scheme as \(\frac{(0.2+N/n)}{0.2}\) where \(N\) is the number of vertices and \(n\) is the number of iterations. To find a good initial \(\beta\) and how often to apply the modification, we ran our code many times by taking random samples for the initial \(\beta\) and the \('trials'\)(modifying in every that number of iteration). As a result, we obtained the best results when we used 1 billion iterations with initial \(\beta =0.98\) and \('trials'=3.4*N\). This simulation of 1 billion iterations is completed in just 100 seconds. Also, when we increase the number of iterations, we could not observe a better \(H_{min}\). 
On the other hand, since we are obliged to give the graphs of \(H\) and \(H_{min}\) in the report, we decided to use 1 million iterations for plotting them in Matlab without crashing the program. Therefore, we also changed initial \(\beta =0.8\) and \('trials'=1.5*N\) which we get good results for this number of iterations. The final \(\beta\) was 23 which makes the acceptance probability close enough to zero as to be considered as the chain is frozen. As we expected, results were not as good as it is in 1 billion iterations, but still close to it.

\begin{figure}[t]
    \centering
    \scalebox{0.45}{\includegraphics{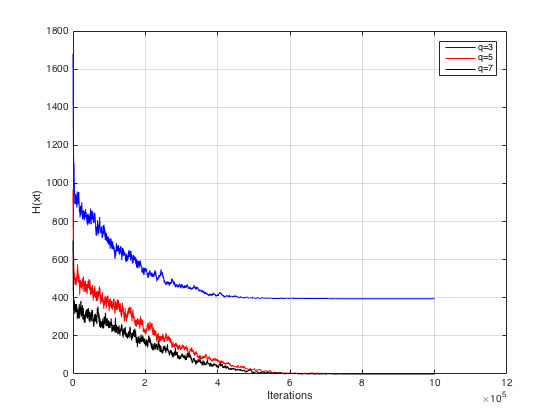}}
    \caption{Results on an Erdos-Renyi graph with $1000$ vertices and average node degree $c=10$. The algorithm is iterated $10^6$ times for every $q$ value of $3$,$5$ and $7$. For these $q$ values, $H_{min}$ is obtained as $395$, $0$,and $0$ respectively.}
\end{figure} 

We plotted the curves of \(H(x^{t})\) as a function of time for three different values of \(c=5, 10, 20\) and three different values of \(q=3, 5, 7\). The plots can be seen in Figures 1,2 and 3.

In general, as we desired, there are significant oscillations in the \(H\) values at the beginning (\(\beta\) is small) of iterations and they become more stable with the time and at the end (\(\beta\) is large) they are frozen.

From the perspective of \(c\) values, the number of edges in the graph is proportional to the \(c\) value. Thus, with higher connectivity, it is harder to obtain a proper coloring scheme for the graph. This can be verified from the Figures 1,2 and 3, because it is seen that when \(c=5\), \(q=5\) and \(q=7\) give proper coloring where \(q=3\) gives a coloring scheme which has a relatively small cost. When \(c=10\), \(q=5\) and \(q=7\) again give proper coloring where \(q=3\) gives a coloring scheme which has a higher cost than \(c=5\) case. Finally, when \(c=20\), only \(q=7\) is able to give a proper coloring, \(q=5\) no longer provides a proper coloring scheme and \(q=3\) has the highest cost among all schemes.

From the perspective of \(q\) values, it is straight-forward to state that when \(q\) increases, we obtain smaller costs in every iteration, and it is intuitively obvious since higher number of available color means higher possibility of having different colors on two ends of an edge. Although there is a possibility that in some iterations we can have smaller cost in colorings while also having smaller \(q\) values, this probability is very small and even gets smaller when the vertex number or iterations (with same modification scheme of \(\beta\)) increases.


\begin{figure}[t]
    \centering
    \scalebox{0.45}{\includegraphics{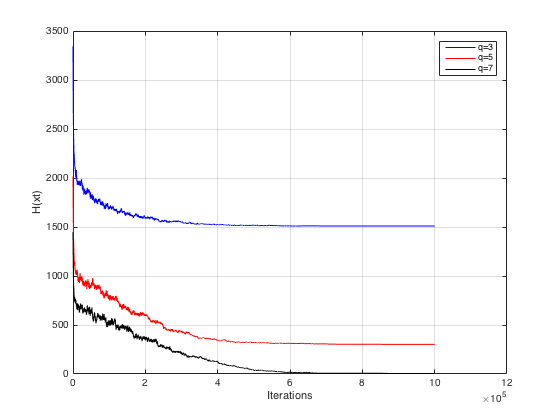}}
    \caption{Results on an Erdos-Renyi graph with $1000$ vertices and average node degree $c=20$. The algorithm is iterated $10^6$ times for every $q$ value of $3$,$5$ and $7$. For these $q$ values, $H_{min}$ is obtained as $1510$, $302$, and $5$ respectively.}
\end{figure} 

In our second experiment, we aim to show the relation between the node degree of Erdos-Renyi graph and the minimum of the cost function (\(H_{min}\)). For the experiment setup, we created 100 random graphs whose \(c\) parameters range from 1 to 100. On each of these graphs, we attempted to get the best coloring for \(q=3,5,7\) using our simulated annealing algorithm implementation. We kept the parameters of the algorithm the same as we set in the first experiment.

As a result, we got the plot in Figure 4. As expected, when the node degree is very low, \(q=3,5,7\) all gives proper coloring for the problem. When we gradually increase the node degree, minimum value of the cost function (\(H_{min}\)) starts to be non-zero for \(q=3\) at first, then for \(q=5\) and lastly for \(q=7\). This is a normal behavior because we can color a graph easier if we have more colors to use, namely we have more chance to get closer to proper coloring by using more colors. Also, after being non-zero, linear increase of
\(H_{min}\) can be seen from the graphs. This can be explained like that: The expectation of edge number will be \(c(N-1)/2\), when the node degree \(c\) is increased linearly, edge number will increase in the same pattern, and as edge number increases and since we have random coloring, we will also have a linear increase in the minimum of the cost function \(H_{min}\). But, the slope of the linearly increasing functions are bigger when \(q\) is smaller. This is because the increase in the total number of possible color combinations will grow more while \(q\) is increasing. For instance, think of a graph which has 3 vertices and all of the vertices are connected. In this case, we have 6 different proper colorings if \(q=3\), 24 if \(q=4\) and 60 if \(q=5\). So, as one can see, we have increasingly growing options while \(q\) value increases which explains the different values of the slopes of the \(H_{min}s\).

\begin{figure}[t]
    \centering
    \scalebox{0.45}{\includegraphics{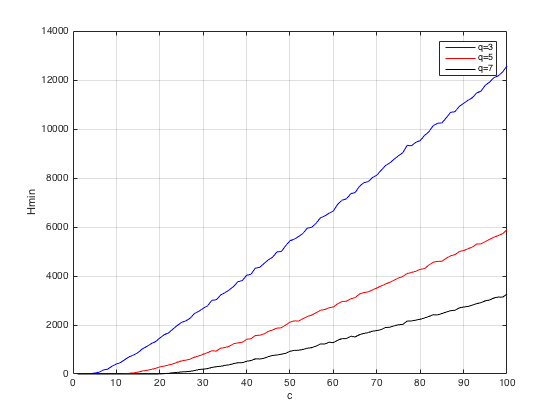}}
    \caption{The plot showing the relation between average node degree ($c$) and the minimum potential. $c$ values ranges from $1-100$ and the algorithm is iterated $10^6$ times for every $q$ value of $3$,$5$ and $7$.}
\end{figure}


\section{Conclusion}

1) The more the iteration the larger the probability of having better coloring.The reason for this is we have the possibility of making more iterations for each \(\beta\) value which makes us closer to mixing time for that \(\beta\) value. 

2) Different \(\beta\) modification schemes give similar results given the required conditions are satisfied (small initial \(\beta\), modifying slow enough not to get stuck in a local minimum).

3) Final \(\beta\) value must be large enough to guarantee that the chain is frozen(1/\(\beta\) goes to 0).

4) The simulated annealing algorithm on large instances of graph coloring problem does not assure the optimal coloring assignment. The mixing time for the chain to converge to the stationary distribution is larger than the time required for exploring the state space of the problem. That state space is quite large, for example, when the number of vertices is 1000 and number of colors are 5, the size is $5^{1000}$. 

5) Since the simulated annealing algorithm is a randomized algorithm and it does not guarantee reaching the optimal value, different runs produce different results. In order to obtain the minimum potential in limited time, the best strategy is to split the time interval into several runs and  
return the minimum value among the runs. Another advantage of this strategy is that different runs can be parallelizable due to independence of the runs. We run the algorithm concurrently on the same graph instance, achieving higher computational power.

6) C++ is efficient in time but harder to code.


\end{document}